\documentclass[sigconf]{acmart}

\AtBeginDocument{%
  }

\setcopyright{none}
\usepackage{graphicx}
\usepackage{textcomp}
\usepackage{xcolor}
\usepackage{balance}
\usepackage{float}
\usepackage{array}
\usepackage{makecell}
\usepackage{booktabs}

\begin{document}

\title{\textsc{HLSClaw}: Bringing Vibe Coding to High-Level Synthesis through Skill-Based Agentic Harness}

%

%

\author{Jinming Ge}
\authornote{Corresponding author: Jinming Ge (jgeab@connect.ust.hk)}
\orcid{0009-0008-7986-4886}
\affiliation{\institution{The Hong Kong University of Science and Technology}
\city{Hong Kong}
\country{Hong Kong}}
\email{jgeab@connect.ust.hk}

\author{Shang Liu}
\orcid{0009-0000-0057-7844}
\affiliation{\institution{The Hong Kong University of Science and Technology}
\city{Hong Kong}
\country{Hong Kong}}
\email{sliudx@connect.ust.hk}

\author{Jiawei Liang}
\orcid{0000-0003-0748-3183}
\affiliation{\institution{The Hong Kong University of Science and Technology}
\city{Hong Kong}
\country{Hong Kong}}
\email{jliangbr@connect.ust.hk}

\author{Shangkun Li}
\orcid{0009-0000-7623-4178}
\affiliation{\institution{The Hong Kong University of Science and Technology}
\city{Hong Kong}
\country{Hong Kong}}
\email{shangkun.li@connect.ust.hk}

\author{Afzal Ahmad}
\orcid{0000-0003-4491-5440}
\affiliation{\institution{The Hong Kong University of Science and Technology}
\city{Hong Kong}
\country{Hong Kong}}
\email{afzal.ahmad@connect.ust.hk}

\author{Jiang Xu}
\orcid{0000-0001-9089-7752}
\affiliation{\institution{The Hong Kong University of Science and Technology (Guangzhou)}
\city{Guangzhou}
\country{China}}
\email{jiang.xu@hkust-gz.edu.cn}

\author{Linfeng Du}
\authornote{Corresponding author: Linfeng Du (linfeng.du@connect.ust.hk)}
\orcid{0000-0002-3007-4890}
\affiliation{\institution{The Hong Kong University of Science and Technology}
\city{Hong Kong}
\country{Hong Kong}}
\email{linfeng.du@connect.ust.hk}

\author{Wei Zhang}
\authornote{Corresponding author: Wei Zhang (wei.zhang@ust.hk)}
\orcid{0000-0002-7622-6714}
\affiliation{
\department{Dept of Electronic \& Computer Engineering}\institution{The Hong Kong University of Science and Technology}
\city{Hong Kong}
\country{Hong Kong}}
\email{wei.zhang@ust.hk}

\renewcommand{\shortauthors}{Ge et al.}

\begin{abstract}

LLM agents are rapidly shifting software development toward intent-driven \emph{vibe coding}, yet their effectiveness drops sharply in high-level synthesis (HLS), where code optimization must jointly consider design structure, directive search, and downstream hardware constraints. Existing HLS optimization approaches, including both heuristic-based methods and recent LLM-assisted pipelines, typically operate at fixed abstraction levels or in weakly coupled stages, limiting their ability to leverage the strong interplay between source code restructuring and directive search outcomes.

Our key insight is that HLS optimization can be reframed from a fixed, hardwired transformation pipeline into a skill-based agentic control problem: an LLM agent equipped with modular, executable optimization skills and a feedback-driven orchestrator can autonomously navigate the coupled space of code restructuring and directive search. This motivates the design of \textsc{HLSClaw}\footnote{HLSClaw is open-sourced at \url{https://github.com/gjskywalker/HLSClaw}}, the first HLS-specific, skill-based agentic harness for holistic HLS optimization. On real-world HLS benchmarks, \textsc{HLSClaw} achieves a $39.12\times$ geometric-mean latency speedup, outperforming the compiler baseline ScaleHLS by $1.98\times$, and surpassing the commercial coding agents Claude Code and Codex by $1.86\times$ and $1.70\times$, respectively, while maintaining strong timing robustness.
\end{abstract}





\maketitle





\section{Introduction}
\label{sec:introduction}

Software development is undergoing a paradigm shift from labor-intensive manual coding toward AI-assisted \textit{vibe coding}, in which developers express high-level intent to LLM-powered agents. This paradigm succeeds in mainstream software since modern software compilers are mature enough to accommodate structural variations in source code without materially affecting runtime correctness or performance~\cite{ashouri2018survey}. General-purpose agentic tools such as OpenAI Codex~\cite{openai2025codex} and Claude Code~\cite{anthropic2025claudecode} capitalize on this robustness, delivering remarkable proficiency in mainstream software tasks with minimal human intervention.


However, this paradigm breaks down in high-level synthesis (HLS), the dominant agile design flow for specialized hardware accelerators~\cite{cong2022fpga,chi2022democratizing,zhang2025fpt}.
Unlike software compilation, HLS design quality is highly sensitive to both source-level computational structure and applied hardware directives. Their interaction shapes the resulting hardware architecture and quality of results (QoR), giving rise to a multidimensional and tightly coupled optimization space.
Consequently, closing the \emph{vibe coding} quality gap for HLS demands \textbf{domain-specific harnesses}—structured wrappers that equip an LLM agent with executable tools, feedback parsers, and correctness checks—to perform high-quality source-to-source transformations, the decisive factor for final hardware QoR.

\begin{figure}[t]
    \centering
    \includegraphics[width=1\linewidth]{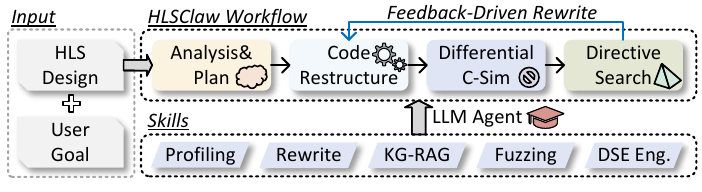}
    \vspace*{-2\baselineskip}
    \caption{Overview of \textsc{HLSClaw}: an LLM agent coordinates optimization skills in a feedback-driven workflow.}
    \vspace*{-1.5\baselineskip}
    \label{fig:framework}
    \Description{An HLS design and a user goal enter a workflow of analysis and planning, code restructuring, differential C simulation, and directive search. A feedback arrow returns from directive search to code restructuring. An LLM agent orchestrates five underlying skills: Profiling, Rewrite, KG-RAG, Fuzzing, and the DSE Engine.}
\end{figure}

\begin{figure*}[t]
    \centering
    \includegraphics[width=0.93\linewidth]{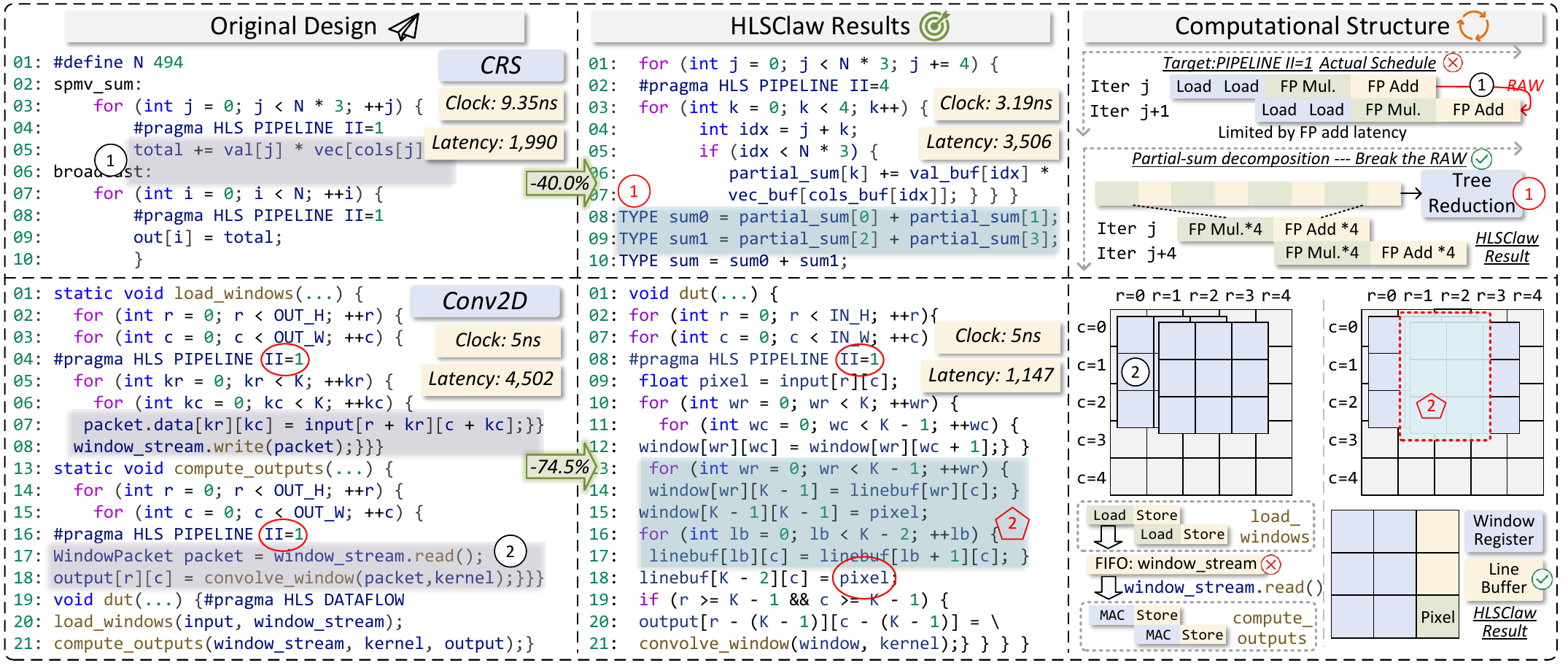}
    \vspace*{-1\baselineskip}
    \caption{Motivation examples demonstrating the limitations of existing HLS compilers and the opportunity for LLMs. The green arrows show execution time reductions. Circle and pentagon markers highlight the bottlenecks and modified regions.}
    \label{fig:motivation}
    \Description{Two rows compare original code, HLSClaw rewrites, and the corresponding hardware structures. For CRS, a loop-carried floating-point accumulation dependency is replaced by four partial sums and a tree reduction. Latency increases from 1,990 to 3,506 cycles, while the clock period decreases from 9.35 to 3.19 nanoseconds, reducing execution time by 40.0 percent. For Conv2D, repeated loading of overlapping windows through a FIFO is replaced by a line buffer and sliding-window registers. At a 5-nanosecond clock period, latency decreases from 4,502 to 1,147 cycles, reducing execution time by 74.5 percent. Numbered circle and pentagon markers connect the bottlenecks with the rewritten code and structures.}
    \vspace*{-1\baselineskip}
\end{figure*}

Despite sustained efforts from both academia and industry~\cite{cong2016source,ye2022scalehls,cheng2024seer,du2023fado,du2024fado,sohrabizadeh2022autodse,pouget2025unified,xiong2024hlspilot,xu2025hlsrewriter}, high-quality HLS source-to-source transformation remains an open challenge. Existing approaches either isolate code restructuring from HLS directive search or couple them in a fixed order, where early structural transformations constrain the space for downstream directive optimization.

Recent LLM-based methods~\cite{xiong2024hlspilot,xu2025hlsrewriter,collini2025c2hlsc,li2025chathls} alleviate this rigidity by leveraging LLMs' natural capacity for flexible source-level reasoning. However, they still organize optimization as either a linear pipeline or a human-scripted iteration, lacking a closed-loop architecture in which structural decisions can be revised based on later synthesis outcomes. These gaps cannot be addressed by simply prompting the LLM to ``try again'' with raw feedback: without structured feedback parsing, the agent cannot diagnose bottlenecks; 
without tool-driven validation, it silently introduces bugs; and without optimization memory, it repeats failed transformations.

Our key insight is that HLS experts navigate this tightly coupled optimization problem not by following a fixed recipe, but by iteratively diagnosing bottlenecks from synthesis feedback, selecting matched strategies, validating functional consistency, and revising upon deficiency---precisely the closed-loop capabilities that current LLM-based methods lack.
Our work, as shown in Fig.~\ref{fig:framework}, encodes this expert workflow by reframing HLS source-to-source transformation from hardwired rule design to dynamic skill orchestration---packaging expert HLS knowledge into modular skills that an LLM agent selects and composes according to the current design state and vendor-tool feedback. The main contributions are as follows:

\begin{itemize} 
\item We present \textsc{HLSClaw}, the first HLS-specific, skill-based agentic harness that brings vibe coding to HLS through feedback-driven orchestration of code restructuring and directive search. 


\item We equip \textsc{HLSClaw} with modular, composable skills whose complementary benefits for optimization quality and efficiency are demonstrated through ablation studies.




\item Across 12 HLS benchmarks, we achieve a 39.12$\times$ geometric-mean latency speedup, outperforming Claude Code by 1.86$\times$ and Codex by 1.70$\times$, with only modest resource overhead.

\end{itemize}



\section{Motivation}
\label{sec:motivation}

We present two representative design snippets to illustrate the fundamental limitations of current HLS optimization: the inability of HLS compilers to compensate for suboptimal source structure, and the necessity of bidirectional coupling between code restructuring and directive search for high-quality optimization. For clarity, Fig.~\ref{fig:motivation} only retains the key computational structure of each kernel. All synthesis results target the AMD U250 FPGA at a 5 ns clock period (200 MHz) using Vitis HLS 2025.2.

\subsection{When Directives Are Not Enough}
\label{sec:crs}
Fig.~\ref{fig:motivation} (top) presents the CRS kernel, a sparse reduction-and-broadcast pattern that commonly appears in ML workloads. The original design uses the aggressive directive \texttt{\#pragma HLS PIPELINE II=1}~(\raisebox{-2.3662pt}{\includegraphics[width=9.03387pt]{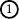}}). This setting yields a latency of 1{,}990 cycles. However, it ignores a loop-carried RAW dependency on \texttt{total}: the requested \texttt{II=1} is unachievable in practice due to the FP accumulation latency, and the resulting design faces severe timing pressure with the Vivado post-synthesis clock stretched to 9.35\,ns. The indirect memory access \texttt{vec[cols[j]]} further complicates scheduling since memory partitioning cannot guarantee conflict-free bank access. As a result, although the transformed code appears to reduce cycle count, the generated hardware is timing-infeasible and yields worse QoR.

By contrast, \textsc{HLSClaw} identifies the loop-carried RAW dependency as the key bottleneck and applies a coordinated restructuring (\raisebox{-2.3662pt}{\includegraphics[width=9.03387pt]{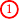}}): it stages external arrays into BRAM, decomposes the accumulator into \texttt{UNROLL\ FACTOR=4} independent partial sums with the inner loop fully unrolled, and relaxes the pipeline target to \texttt{II=4} so each \texttt{partial\_sum[k]} is updated once every four iterations. A final tree reduction merges the partial results, breaking the RAW hazard and producing a more efficient hardware pipeline. Although this rewrite results in a higher latency of 3{,}506 cycles, it achieves a much shorter clock of 3.19\,ns, leading to a lower execution time. 




\subsection{When One Strategy Does Not Fit All}
\label{sec:conv2d}

The $3 \times 3$ convolution over a $32 \times 32$ input shown in Fig.~\ref{fig:motivation}~(bottom) exposes a limitation of task-level dataflow decomposition~\cite{ye2022scalehls,basalama2025stream,pouget2025holistic,abi2026fifoadvisor}.
After applying dataflow decomposition with pipeline directives at $\mathrm{II}=1$, the resulting latency remains far from optimal despite significant engineering effort.
The computational structure exposes the underlying cause: nine input values are loaded independently for every output pixel, yet adjacent windows overlap by six elements  (\raisebox{-2.3662pt}{\includegraphics[width=9.03387pt]{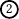}}).
The true bottleneck is \emph{spatial data reuse}, and factoring the code into producer--consumer stages does nothing to exploit it.

\textsc{HLSClaw} addresses this bottleneck by reorganizing the entire loop nest around a line-buffer and sliding-window architecture (\raisebox{-2.49675pt}{\includegraphics[width=10.28084pt]{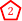}}).
A single pipelined loop at \texttt{II\,=\,1} brings in one new pixel per cycle, shifts a $K \times K$ window of registers, and refills one column from the line buffer, with the MAC fully unrolled.
The resulting latency of 1{,}147 cycles ($-74.5\%$ vs.\ original) demonstrates the payoff of matching the transformation to the kernel-specific bottleneck.
Crucially, this improvement stems from directive search outcomes feeding back into restructuring decisions: when the dataflow-decomposed code proved unresponsive to further directive tuning, \textsc{HLSClaw} recognized diminishing returns and selected a fundamentally different restructuring strategy---something forward-only pipelines cannot do, as they lack feedback between stages.

\subsection{From Compiler Limitations to LLM Opportunities}
\label{sec:motivation_summary}


These two cases expose two fundamental challenges in existing HLS optimization methods. 
\textbf{Challenge 1: structure-constrained directive exploration.} CRS shows that restrictive computational dependencies fundamentally limit directive search, often trapping it in locally optimal designs. 
\textbf{Challenge 2: missing architecture-level reorganization.} Conv2D shows that even after dataflow decomposition and
pipelining, existing compilers may fail to identify transformations such as line-buffer-based spatial reuse that eliminate the dominant memory-access bottleneck.

Both challenges arise because the required restructurings are high-level, domain-aware optimizations that are difficult to encode as deterministic compiler transformations, motivating the use of LLM-based reasoning. 
However, existing LLM-assisted HLS methods typically lack a feedback connection between code restructuring and downstream DSE, preventing synthesis outcomes from revising earlier structural decisions. 
HLSClaw addresses this limitation by grounding the agent with domain-specific skills, coordinating them in a feedback-driven loop, and using context engineering to preserve diagnoses, tool evidence, and design history across iterations. This design harnesses the flexibility of LLMs to enable a tool-grounded, correctness-aware, and iterative process tailored to the tightly coupled nature of HLS optimization.

\section{Methodology}
\label{sec:methodology}

\subsection{Framework Overview}
\label{sec:overview}

\begin{figure}[t]
    \centering
    \includegraphics[width=0.9\linewidth]{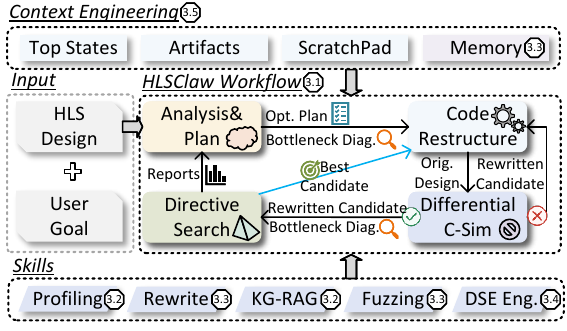}
    \vspace*{-1\baselineskip}
    \caption{Overview of \textsc{HLSClaw} with five optimization skills, a feedback-driven workflow, and context engineering.}
    \vspace*{-1\baselineskip}
    \label{fig:architecture}
    \Description{The architecture combines a context-management layer, a feedback-driven optimization workflow, and five executable skills. An input HLS design and user goal guide analysis and planning, which sends an optimization plan and bottleneck diagnosis to code restructuring. Differential C simulation compares the original design and rewritten candidate; failures return to restructuring, while passing candidates proceed to directive search. Search reports return to analysis, and the best candidate feeds subsequent restructuring. Top States, Artifacts, ScratchPad, and Memory support the workflow, with Profiling, Rewrite, KG-RAG, Fuzzing, and DSE Engine skills below.}
\end{figure}

Fig.~\ref{fig:architecture} presents the architecture of \textsc{HLSClaw}. 
Given an HLS design and a user-specified optimization goal, the agent first invokes the \emph{Profiling} and \emph{KG-RAG} skills to analyze synthesis feedback, diagnose the bottleneck, and formulate an optimization plan. 
Guided by this plan, the agent invokes the \emph{Rewrite} skill to generate source-level design variants that target the identified bottleneck.
The rewritten candidate is then compared with the original design through differential C simulation using the \emph{Fuzzing} skill. 
Candidates that fail the functional-consistency check are rejected and returned to the agent for further revision, whereas
valid candidates proceed to directive exploration. 
The agent then invokes the \emph{DSE Engine} skill to explore directive configurations for the current design and evaluates them through HLS synthesis.
After directive search, the resulting QoR reports are fed back to the analysis stage. 
If the best candidate violates the user-specified constraints or the reports indicate further optimization opportunities, the agent updates its bottleneck diagnosis and initiates another restructuring iteration. 
This feedback-driven process continues until no candidate provides a meaningful QoR improvement or the optimization budget is exhausted.  

\subsection{Profiling \& Knowledge-Guided Planning}
\label{sec:profiling&KG-RAG}

A key enabler of \textsc{HLSClaw}'s agentic behavior is its ability to \emph{plan before acting}. 
Before applying any optimization, the agent invokes the
\emph{Profiling} skill, which runs HLS synthesis and performs static analysis of the resulting QoR reports to derive higher-level bottleneck diagnoses.
Crucially, profiling is not limited to the initial design: after major rewrites or DSE rounds, the agent re-invokes the \emph{profiling} skill to reassess the updated design. 
This ensures that subsequent decisions are grounded in the current design state and forms a central component of \textsc{HLSClaw}'s closed-loop optimization. 

\begin{figure}[h]
    \centering
    \includegraphics[width=0.9\linewidth]{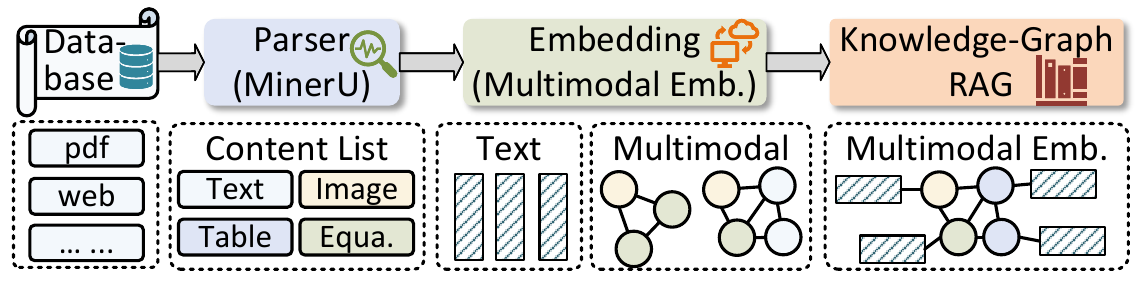}
    \vspace*{-1\baselineskip}
    \caption{Offline construction of the KG-RAG library from multimodal documents for HLS knowledge retrieval.}
    \vspace*{-1\baselineskip}
    \label{fig:kgrag}
    \Description{A left-to-right pipeline constructs a retrieval library from PDF files, web pages, and other documents. MinerU parses each document into text, images, tables, and equations. Multimodal embedding encodes the extracted content. The resulting knowledge graph links multimodal content and text representations for retrieval-augmented generation.}
\end{figure}

After obtaining the QoR report, the agent invokes the \emph{KG-RAG} skill to ground its bottleneck diagnosis in
domain-specific HLS knowledge. 
As shown in Fig.~\ref{fig:kgrag}, the knowledge library is constructed offline by parsing vendor documentation with a multimodal document parser~\cite{guo2025rag}, encoding the extracted text, images, tables, and equations, and organizing them into a knowledge-graph-based RAG store. 
At query time, the agent formulates a bottleneck-aware query, and the RAG library returns prioritized optimization strategies with supporting rationale. 
Unlike Flat-RAG, which retrieves semantically similar passages independently, the graph-structured RAG explicitly models relationships among bottlenecks, root causes, optimization techniques, and vendor constraints. 
This structure supports multi-hop reasoning, composition of compatible strategies, and vendor-specific grounding.

\subsection{Code Restructuring \& Differential C-Sim}
\label{sec:rewrite}

The \emph{Rewrite} skill provides a library of more than 100 curated code templates collected from publicly available repositories and programming platforms. 
Additionally, we classify them according to common HLS bottlenecks, including redundant computation, loop-carried dependencies, inefficient memory access, limited data reuse, irregular control flow, and imbalanced dataflow organization.
For each bottleneck category, the library records representative code templates together with their applicable conditions, enabling the agent to retrieve structurally relevant examples when implementing an optimization plan. 
The template library is further complemented by a self-updating memory. 
When the agent encounters a bottleneck that is not adequately covered by existing templates, 
it derives a new rewrite pattern from the successful optimization attempt and stores the validated pattern, its triggering
bottleneck, and the observed synthesis outcome for reuse in subsequent runs.

Since LLM-generated rewrites may introduce functional errors, \textsc{HLSClaw} invokes the \emph{Fuzzing} skill before each rewritten candidate enters HLS evaluation. 
The skill constructs a self-checking differential testbench that executes the original design and the rewritten candidate on identical fixed-seed, fuzz-generated inputs through AMD Vitis C simulation. 
Their outputs are compared element-wise using exact matching for integer and fixed-point computations and a predefined numerical tolerance for floating-point outputs. 
A candidate is rejected if it produces an output mismatch, runtime failure, assertion violation, or incompatible top-level interface. 
The failing input and first divergent output are summarized as feedback for the next rewrite attempt. Candidates that pass this test-based functional-consistency check proceed to HLS synthesis and directive search. 
This lightweight gate prevents incorrect rewrites from consuming expensive HLS evaluations.

\begin{figure}[htpb]
    \centering
    \includegraphics[width=0.9\linewidth]{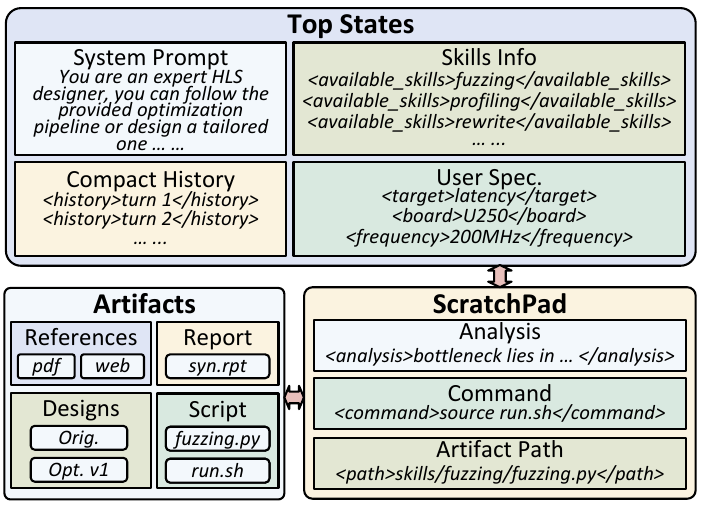}
    \vspace*{-1\baselineskip}
    \caption{Context engineering with Top States, ScratchPad, and Artifacts for orchestration, memory, and outputs.}
    \vspace*{-1\baselineskip}
    \label{fig:context_engineering}
    \Description{Three connected context layers organize an agent run. Top States holds the system prompt, available-skill information, compact interaction history, and user specifications such as the latency objective, U250 board, and 200-megahertz target. ScratchPad holds the current analysis, commands, and artifact paths and exchanges information with Top States. Artifacts stores reference documents, synthesis reports, original and optimized designs, and scripts, and exchanges information with ScratchPad.}
\end{figure}

\subsection{Directive Search \& DSE Engine}
\label{sec:pragma_dse}

The directive-search stage consists of two steps. 
First, the agent leverages the structural analysis from the rewrite stage to identify high-impact directive sites and construct a focused candidate set from the full set of AMD Vitis HLS directives. 
The agent then invokes the \emph{DSE Engine} skill to evaluate these candidates through an algorithmic search engine, progressively adding or updating one directive site at each iteration. 

For candidate selection, HLSClaw adopts a constraint-first, goal-aware ranking scheme. 
Let $\mathcal{C}$ denote the set of user-specified constraints, where each constraint is expressed as $g_c(x)\leq b_c$ for candidate $x$. 
These constraints may include the target clock period and resource budgets. 
We quantify the maximum normalized constraint violation as
\begin{equation}
\label{eq:violation}
V(x)=
\max_{c\in\mathcal{C}}
\left[
\frac{g_c(x)-b_c}{b_c}
\right]_{+},
\qquad
[z]_{+}=\max(z,0).
\end{equation}
A candidate is feasible if and only if $V(x)=0$.

Candidates are first ranked by normalized constraint violation and then by the user-selected optimization objective.
Let $q_g(x)$ denote the QoR cost of candidate $x$ under goal $g$. 
We normalize this objective against the current rewritten baseline $x_0$:
\begin{equation}
\label{eq:objective}
J_g(x)=\frac{q_g(x)}{q_g(x_0)}.
\end{equation}
A lower value of $J_g(x)$ indicates a better design under the selected
optimization goal. 
Candidates are ordered lexicographically by
\begin{equation}
\label{eq:ranking}
\Phi_g(x)=\bigl(V(x),J_g(x)\bigr).
\end{equation}
Each feasible candidate is preferred over an infeasible one; feasible candidates are then ranked solely by the selected QoR objective.

The \emph{DSE Engine} explores hardware directives in three stages.
It first synthesizes the rewritten baseline and evaluates all single-site directive candidates in parallel. 
For each site, the best option that improves upon the baseline under the lexicographic ranking $\Phi_g=(V,J_g)$ is retained. 
These options are then progressively combined through beam expansion. 
At each round, the engine ranks the evaluated configurations using $\Phi_g$ and applies rank-based Boltzmann sampling to preserve promising alternatives while avoiding early commitment to a local optimum. 
As the search progresses, the selection gradually favors higher-ranked candidates.
Finally, the remaining evaluation budget is used to explore unseen, conflict-free multi-site combinations, and the lexicographically best configuration is returned.

 \begin{figure*}[t]
      \centering

      \includegraphics[width=0.9\textwidth]
      {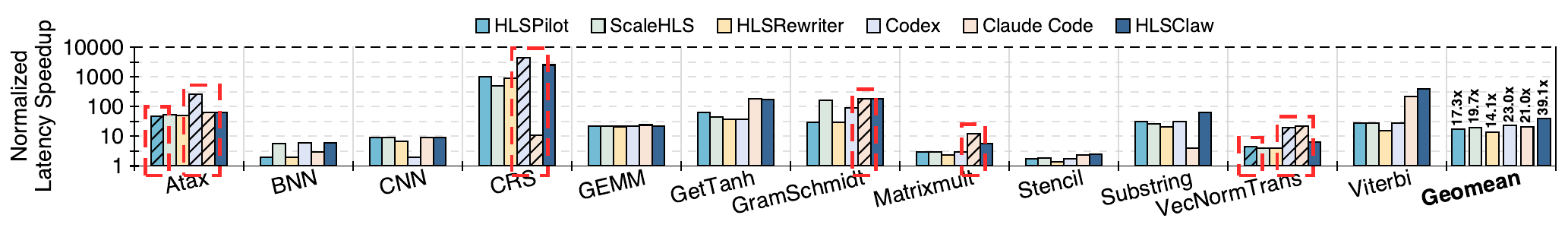}
      \par\vspace{-0.7em}

      \includegraphics[width=0.9\textwidth]
      {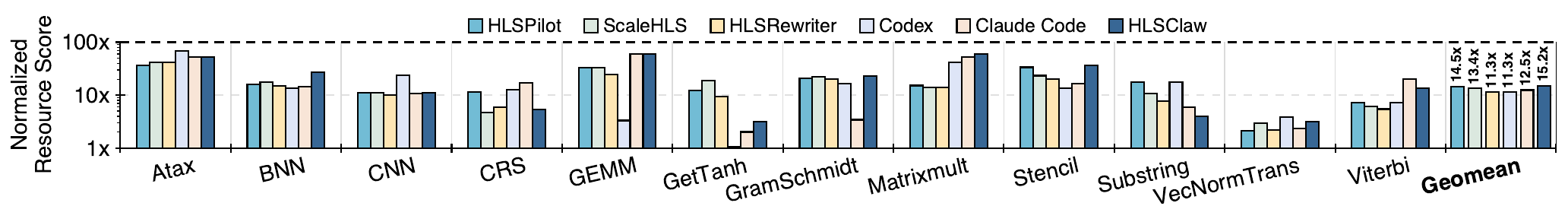}
      \vspace*{-1.3\baselineskip}
      \caption{
          Performance and resource comparison across 12 HLS benchmarks.
          Top: normalized latency speedup, where higher is better.
          Bottom: normalized resource score relative to the original design, where lower is better.
      }
      \label{fig:performance-resource-comparison}
      \Description{Two grouped bar charts compare HLSPilot, ScaleHLS, HLSRewriter, Codex, Claude Code, and HLSClaw on Atax, BNN, CNN, CRS, GEMM, GeTanh, GramSchmidt, Matrixmult, Stencil, Substring, VecNormTrans, and Viterbi, followed by geometric means. Both vertical axes are logarithmic. The upper chart shows normalized latency speedup, where higher is better; geometric means in method order are 17.3, 19.7, 14.1, 23.0, 21.0, and 39.1 times. The lower chart shows normalized resource score, where lower is better; geometric means are 14.5, 13.4, 11.3, 21.3, 12.5, and 15.2 times. HLSClaw has the highest geometric-mean speedup, with resource use varying across kernels.}
      \vspace*{-1\baselineskip}
  \end{figure*}

\subsection{Context Engineering}
\label{sec:context_engineering}


Long-horizon agentic runs accumulate heterogeneous context, including user constraints, intermediate analyses, code variants, tool outputs, and empirical takeaways. As illustrated in Fig.~\ref{fig:context_engineering}, \textsc{HLSClaw} partitions this context into three layers. \emph{Top States} maintains stable orchestration information such as the system prompt, skill descriptions, execution history, and user specifications. \emph{ScratchPad} stores transient reasoning state, including bottleneck diagnoses, commands, and artifact paths, while \emph{Artifacts} preserves concrete outputs such as designs, synthesis reports, and scripts. This separation keeps control signals explicit, allows reasoning to persist across skill invocations, and preserves inspectable evidence throughout long-horizon optimization.


This partitioning yields three benefits. First, \emph{robustness}: the agent can resume, reroute, or terminate using explicit state rather than fragile conversational reconstruction. Second, \emph{reasoning quality}: separating working memory from orchestration metadata reduces interference across skill
invocations. Third, \emph{auditability}: tool outputs persist as artifacts rather than free-form summaries. Context engineering therefore provides a reliable basis for multi-skill orchestration by separating control logic, evolving reasoning state, and verifiable optimization evidence.



\label{sec:experiment}
  
\section{Experiments}

We implement \textsc{HLSClaw} using LangGraph 1.0.10~\cite{langgraph2024}, a production-grade framework for building agentic systems in industry~\cite{bytedance2025deerflow,langchain2024top5}, comprising over 12{,}000 lines of Python code, and evaluate it on a suite of 12 representative HLS kernels drawn from four classical benchmark suites~\cite{pouchet2012polybench,hara2008chstone,zhou2018rosetta,reagen2014machsuite} spanning dense linear algebra (GEMM, Matrixmult, Atax, GramSchmidt), neural-network-style computation (BNN, CNN, GetTanh), stencil and convolution patterns (Stencil, CRS), and control-heavy or irregular workloads (Viterbi, Substring, VecNormTrans).
All designs are synthesized using Vitis HLS 2025.2 targeting the AMD U250 at a 5\,ns clock period (200\,MHz), with latency as the primary optimization objective. 

For DSE, we set the beam width to $B=5$, limit each search to at most $M=13$ HLS evaluations per rewritten design, and use an initial Boltzmann temperature of $T_0=50.0$ with a geometric cooling factor of $\gamma=0.85$. 
Hardware QoR is measured from synthesis reports in terms of latency (cycles), timing slack (ns), and resource utilization. 
We quantify overall hardware cost using a weighted normalized resource
metric, denoted as ResourceScore:
$\mathrm{ResourceScore}
=0.30u_{\mathrm{LUT}}+0.10u_{\mathrm{FF}}
+0.30u_{\mathrm{DSP}}+0.30u_{\mathrm{BRAM}}$.
This formulation assigns larger weights to LUT, DSP, and BRAM because they are typically more constraining in HLS designs, while FF receives a smaller weight due to its lower practical pressure.

We compare against five baselines representing three optimization paradigms: (1)~\emph{compiler-based}: ScaleHLS~\cite{ye2022scalehls}; (2)~\emph{LLM-based}: HLSPilot~\cite{xiong2024hlspilot} and HLSRewriter~\cite{xu2025hlsrewriter}; (3)~\emph{general-purpose coding agents}: OpenAI Codex~\cite{openai2025codex} and Claude Code~\cite{anthropic2025claudecode}. To ensure fair comparison across LLM-based methods, \textsc{HLSClaw}, HLSPilot~\cite{xiong2024hlspilot}, and HLSRewriter~\cite{xu2025hlsrewriter} all use GPT-5.5 as the base model.

\subsection{Comparison with SOTA HLS Opt. Methods}
\label{sec:exp_speedup}

Fig.~\ref{fig:performance-resource-comparison} compares end-to-end speedup and resource usage across 12 benchmarks for six methods. 
\textsc{HLSClaw} achieves the highest geometric-mean latency speedup of $39.1\times$, outperforming Codex ($23.0\times$) and Claude Code ($21.0\times$) by $1.70\times$ and $1.86\times$, respectively.
Although the coding agents report higher nominal speedups on several benchmarks, four of these designs suffer severe timing violations, as indicated by the red dashed boxes. 
In contrast, \textsc{HLSClaw} maintains timing-feasible performance across the benchmark suite while remaining competitive in resource usage, with a geometric-mean normalized resource score of $15.2\times$.

For LLM-based methods and general-purpose coding agents, the overall runtime is dominated by repeated Vitis HLS evaluations. 
\textsc{HLSClaw} reduces this cost through differential C simulation, which rejects incorrect rewrites before they enter synthesis and avoids unnecessary tool invocations. 
Compiler-based methods such as ScaleHLS instead use analytical QoR models during DSE, trading synthesis-level evaluation fidelity for lower exploration runtime.

\begin{table}[!t]
\centering
\scriptsize
\setlength{\tabcolsep}{3.5pt}
\caption{Execution time $=$ latency $\times$ achieved clock period;
For all four metrics, higher is better.}
\vspace*{-0.8\baselineskip}
\label{tab:metrics}

\resizebox{\columnwidth}{!}{%
\begin{tabular}{lrrrrrr}
\toprule
Metric &
HLSPilot &
ScaleHLS &
\makecell{HLS\\Rewriter} &
Codex &
\makecell{Claude\\Code} &
\textbf{HLSClaw} \\
\midrule
GM
& 11.89
& 18.29
& 14.16
& 13.34
& 14.68
& \textbf{30.59} \\

ETS
& 3.44
& 6.35
& 5.35
& 4.13
& 5.26
& \textbf{7.48} \\

EWS
& 4.41
& 4.46
& 4.99
& 4.42
& 4.68
& \textbf{5.70} \\

Perf/HW Gain
& 0.82
& 1.36
& 1.25
& 1.18
& 1.17
& \textbf{2.01} \\
\bottomrule
\end{tabular}%
}
\parbox{\columnwidth}{\scriptsize
\textbf{GM}: geometric mean of execution-time speedups.\\
\textbf{ETS/EWS}: equal-time and equal-work harmonic-mean speedups,
respectively~\cite{eeckhout2024geomean}.\\
\textbf{Perf/HW gain}: geometric mean of per-benchmark execution-time
speedup divided by normalized ResourceScore overhead.}

\vspace*{-1\baselineskip}
\end{table}

To account for timing infeasibility and enable a fair comparison of performance and resource efficiency, we further evaluate all methods using four execution-time-based metrics, as summarized in Table~\ref{tab:metrics}. 
Execution time is computed as latency multiplied by the achieved clock period, thereby penalizing designs with poor timing closure. 
\textsc{HLSClaw} consistently achieves the best results across all four metrics, attaining a GM speedup of 30.59$\times$, ETS and EWS speedups of 7.48$\times$ and 5.70$\times$, respectively, and the highest Perf/HW gain of 2.01.

\subsection{Skill Ablation Study}

To isolate the contributions of key components of \textsc{HLSClaw}, we conduct three ablation studies: KG-RAG versus Flat-RAG and No-RAG; differential C simulation for early rejection of incorrect rewrites; and isolated, one-shot, and feedback-driven coordination of code restructuring and directive search.

\begin{figure}[!t]
\centering
\includegraphics[width=0.9\columnwidth]{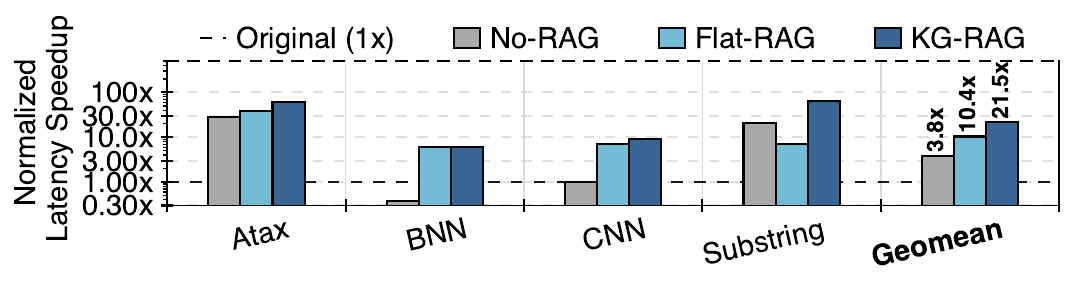}
\par\vspace{-0.56em}
\includegraphics[width=0.9\columnwidth]{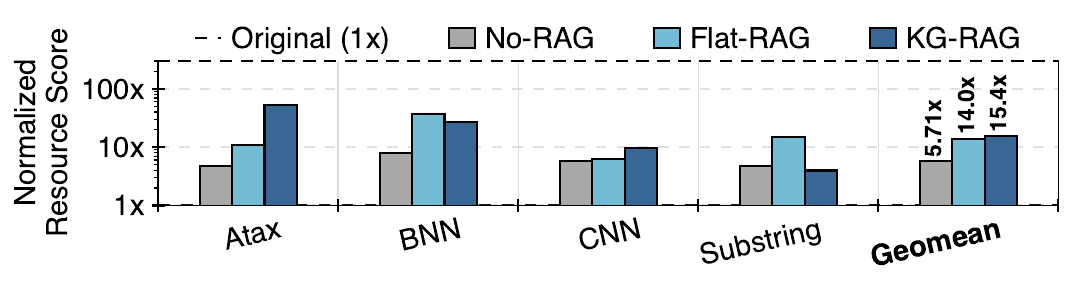}
\vspace*{-1.5\baselineskip}
\caption{KG-RAG ablation on four kernels: (top) normalized latency speedup; (bottom) normalized resource score.}
\label{fig:kgrag_ablation}
\Description{Two grouped bar charts compare No-RAG, Flat-RAG, and KG-RAG on Atax, BNN, CNN, and Substring, followed by geometric means. The logarithmic axes show latency speedup above and normalized resource score below, with the original design marked at one times. Geometric-mean speedups are 3.8, 10.4, and 21.5 times, respectively; corresponding resource scores are 5.71, 14.0, and 15.4 times. KG-RAG provides the highest geometric-mean speedup and speedup per resource score, while resource use varies by kernel.}
\end{figure}

\subsubsection{KG-RAG} 

We evaluate No-RAG, Flat-RAG, and KG-RAG within \textsc{HLSClaw}, as shown in Fig.~\ref{fig:kgrag_ablation}.
KG-RAG achieves the highest geometric-mean latency speedup and Perf/HW gain, with $21.5/15.4 \approx 1.40\times$, compared with
$10.4/14.0 \approx 0.74\times$ for Flat-RAG and $3.8/5.71 \approx 0.67\times$ for No-RAG. 
The small gain of Flat-RAG over No-RAG suggests that isolated passage retrieval provides limited benefit, whereas graph-structured retrieval better captures relationships among bottlenecks, optimization strategies, and hardware constraints.

\subsubsection{Differential C-Sim}

Table~\ref{tab:validation_runtime} evaluates differential C simulation on four benchmarks using four rewrites each. Without early validation, the 16 candidates require $16\times5=80$ HLS evaluations. Eight fixed-seed fuzzing trials reject three incorrect rewrites before DSE, avoiding 15 HLS evaluations. Including 94.10\,s of C-sim overhead, total wall-clock time decreases from 966.50\,s to 389.81\,s.

\begin{table}[t]
  \centering
  \footnotesize
  \setlength{\tabcolsep}{3.5pt}
  \caption{Impact of differential C simulation on validation and
  optimization wall-clock time.}
  \vspace*{-1\baselineskip}
  \label{tab:validation_runtime}
  \resizebox{\columnwidth}{!}{%
    \begin{tabular}{lrrrrr}
      \toprule
      Configuration &
      \makecell{C-sim\\time (s)} &
      \makecell{HLS\\evals} &
      \makecell{Invalid evals\\executed} &
      \makecell{HLS wall\\time (s)} &
      \makecell{Total wall\\time (s)} \\
      \midrule

      Without C-sim gate &
      0 &
      80 &
      15 &
      966.50 &
      966.50 \\

      With differential C-sim &
      94.10 &
      \textbf{65} &
      \textbf{0} &
      \textbf{295.71} &
      \textbf{389.81} \\

      \midrule
      Change &
      +94.10 &
      $-15$ &
      $-15$ &
      $-670.79$ &
      $-576.69$ \\

      Relative saving &
      -- &
      18.8\% &
      100\% &
      69.40\% &
      \textbf{59.70\%} \\
      \bottomrule
    \end{tabular}%
  }
  \vspace*{-1\baselineskip}
\end{table}

\begin{table}[H]
\centering
\scriptsize
\setlength{\tabcolsep}{3pt}
\caption{Contribution of eval-1, eval-2, eval-3, and eval-4.}
\vspace*{-1\baselineskip}
\label{tab:stage}
\resizebox{\columnwidth}{!}{%
\begin{tabular}{lrr@{\hspace{8pt}\rule[-2pt]{0pt}{10pt}}
                    rr@{\hspace{8pt}\rule[-2pt]{0pt}{10pt}}
                    rr@{\hspace{8pt}\rule[-2pt]{0pt}{10pt}}rr}
\toprule
Benchmark &
\makecell{eval-1\\cycles} & \makecell{eval-1\\gain} &
\makecell{eval-2\\cycles} & \makecell{eval-2\\gain} &
\makecell{eval-3\\cycles} & \makecell{eval-3\\gain} &
\makecell{eval-4\\cycles} & \makecell{eval-4\\gain} \\
\midrule
Substring &
24{,}607 & 10.65$\times$ &
42{,}762 & 6.13$\times$ &
4{,}122\textsuperscript{$\dagger$} &
63.59$\times$\textsuperscript{$\dagger$} &
4{,}121 & 63.61$\times$ \\

VecNormTrans &
30{,}729 & 1.46$\times$ &
6{,}199 & 7.26$\times$ &
4{,}814 & 9.35$\times$ &
3{,}127 & 14.39$\times$ \\

CNN &
8{,}115 & 4.61$\times$ &
14{,}278 & 2.62$\times$ &
8{,}115 & 4.61$\times$ &
4{,}105 & 9.11$\times$ \\

Atax &
1{,}259 & 26.62$\times$ &
977 & 34.30$\times$ &
1{,}072\textsuperscript{$\dagger$} &
31.26$\times$\textsuperscript{$\dagger$} &
532 & 63.00$\times$ \\
\bottomrule
\end{tabular}%
}
\vspace{1pt}

\parbox{\columnwidth}{%
\raggedright
\scriptsize
\vspace*{-0.5\baselineskip}
\textsuperscript{$\dagger$} Timing-infeasible design.
}
\vspace*{-2.5\baselineskip}
\end{table}

\subsubsection{Feedback-driven Rewrite}
Table~3 compares directive search only (\emph{eval-1}), code restructuring only (\emph{eval-2}), fixed-order restructuring followed by DSE (\emph{eval-3}), and the full feedback-driven workflow (\emph{eval-4}). The isolated stages favor different kernels, while \emph{eval-3} cannot revise restructuring decisions using downstream synthesis feedback and produces timing-infeasible designs. In contrast, \emph{eval-4} achieves the lowest latency on all four benchmarks while maintaining timing feasibility.


\subsection{Case Study: Multi-Objective Pareto Front}

Although this work mainly focuses on latency optimization, \textsc{HLSClaw} is not restricted to a single optimization objective.
After filtering infeasible candidates, the agent can rank feasible designs using a user-defined latency--resource objective:
\begin{equation}
\label{eq:lat_res_obj}
J_{\lambda}(x)
=
\lambda \frac{L(x)}{L(x_0)}
+
(1-\lambda)\frac{\rho(x)}{\rho(x_0)},
\qquad 0 \leq \lambda \leq 1,
\end{equation}
where $L(x)$ and $\rho(x)$ denote the latency and resource score of candidate $x$, respectively, and $x_0$ is the current rewritten baseline. 

\begin{figure}[t]
\centering
\includegraphics[width=0.85\columnwidth]{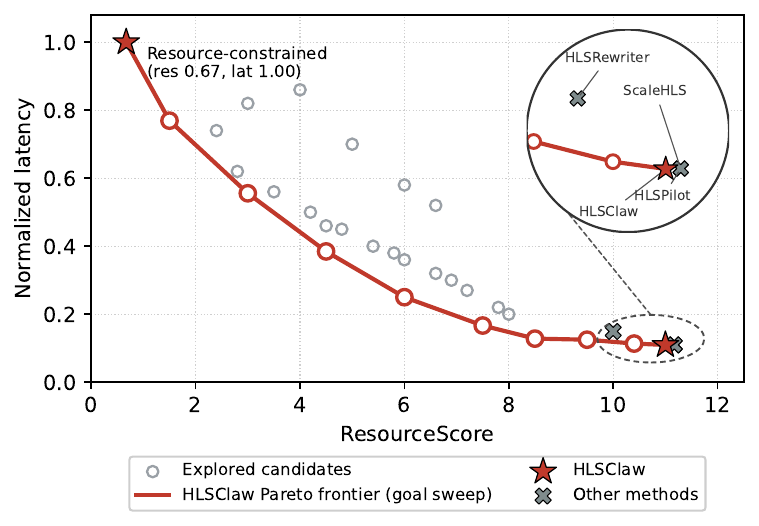}
\vspace*{-1\baselineskip}
\caption{Latency--resource trade-off on CNN. The red curve shows the Pareto frontier obtained by sweeping $\lambda$.}
\label{fig:objsens}
\Description{A scatter plot shows the CNN trade-off between resource score on the horizontal axis and normalized latency on the vertical axis. Gray circles represent explored candidates, and a descending red curve connects the HLSClaw Pareto frontier. A star marks the resource-constrained endpoint at resource score 0.67 and normalized latency 1.00; another star marks a low-latency HLSClaw solution near resource score 11. A magnified inset compares this low-latency region with cross-shaped markers for HLSRewriter, HLSPilot, and ScaleHLS. Moving along the frontier reduces latency at the cost of greater resource use.}
\vspace*{-1.2\baselineskip}
\end{figure}

We sweep $\lambda$ on CNN to study the latency--resource trade-off. As shown in Fig.~\ref{fig:objsens}, \textsc{HLSClaw} explores a broad design space and traces a Pareto frontier from resource-efficient to
latency-oriented solutions, demonstrating its ability to adapt to different user objectives rather than returning only the fastest design.

\subsection{Open-Source Model Study}
\label{sec:opensource_study}

To investigate the feasibility of deploying open-source LLMs for agentic HLS optimization, we establish a local inference environment on an Ascend server equipped with eight Ascend 910B NPUs, each providing 64 GiB of HBM, and a 192-core Kunpeng 920 CPU. The software stack includes CANN 8.5.2, PyTorch 2.7.1, TorchNPU 2.7.1.post2, and Hugging Face Transformers 4.53.3. All evaluated models are executed in BF16 precision.

We evaluate Qwen2.5~\cite{hui2024qwen2} and openPangu~\cite{chen2025pangu} models to validate inference across different model sizes and NPU configurations. For integration with HLSClaw, we deploy openPangu-Embedded-7B on a single NPU and expose it through a lightweight FastAPI server implementing OpenAI-compatible model discovery and chat completion endpoints. HLSClaw communicates with the remote inference server through an SSH tunnel using its existing custom LLM provider, without modifying the agent workflow. The serving configuration uses deterministic decoding (temperature = 0), a maximum generation length of 4096 tokens, and a request timeout of 600 seconds.

\subsubsection{NPU Inference Performance}
\label{sec:npu_performance}

We first evaluate the inference capability of the Ascend platform using three open-source LLM configurations. Table~\ref{tab:npu_inference} reports the execution time and generation throughput from our initial inference tests. Qwen2.5-0.5B and openPangu-Embedded-7B are executed on a single NPU, whereas Qwen2.5-32B is distributed across four NPUs using layer-wise model sharding. Specifically, its 64 Transformer layers are evenly partitioned across the four devices.

\begin{table}[t]
    \centering
    \caption{Inference results of open-source LLMs on
    Huawei Ascend 910B1 NPUs.}
    \label{tab:npu_inference}
    \small
    \setlength{\tabcolsep}{4pt}
    \begin{tabular}{lcccc}
        \toprule
        \textbf{Model} &
        \textbf{NPUs} &
        \textbf{Tokens} &
        \textbf{Time (s)} &
        \textbf{Tokens/s} \\
        \midrule
        Qwen2.5-0.5B     & 1 & 93  & 4.57  & 20.35 \\
        openPangu-7B     & 1 & 92  & 7.18  & 12.81 \\
        Qwen2.5-32B      & 4 & 128 & 32.22 & 3.97  \\
        \bottomrule
    \end{tabular}
\end{table}

All three configurations successfully complete autoregressive generation on the Ascend platform. The 32B experiment further demonstrates the feasibility of distributing a large model across multiple NPUs. Since the measurements involve different model sizes and generation workloads, they serve as functional deployment results rather than a controlled comparison of inference efficiency.

\subsubsection{Open-Source Model Comparison}
\label{sec:opensource_comparison}

\begin{figure}[t]
    \centering
    \includegraphics[width=\linewidth]{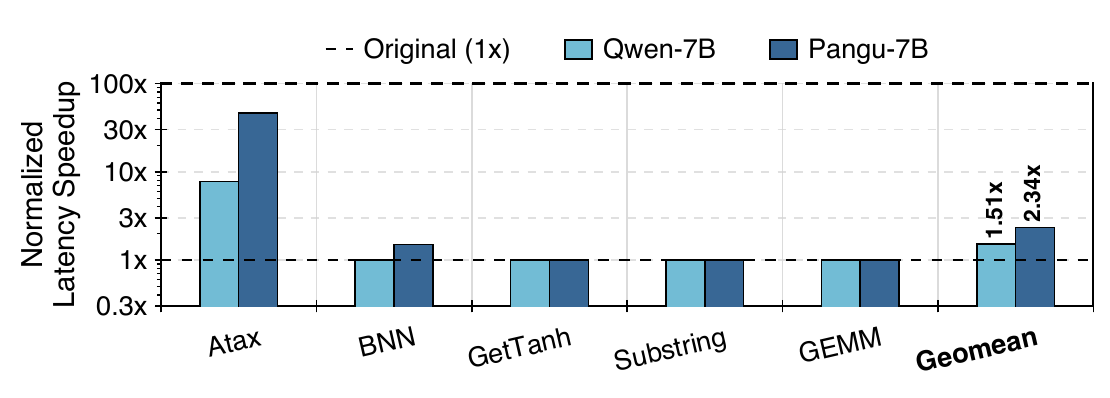}
    \vspace*{-2\baselineskip}
    \caption{Cycle-based latency speedup of HLSClaw using
    openPangu-7B and Qwen2.5-7B-Instruct on five benchmarks.}
    \label{fig:pangu-qwen-7b}
\end{figure}

Figure~\ref{fig:pangu-qwen-7b} compares the HLS optimization performance of openPangu-7B and Qwen2.5-7B-Instruct on five selected benchmarks. Speedup is calculated using worst-case latency in cycles, with unsuccessful runs assigned $1\times$. Pangu achieves a geometric-mean speedup of $2.34\times$,
compared with $1.51\times$ for Qwen, primarily driven by its $46.35\times$ speedup on Atax versus Qwen's $7.77\times$.

However, Pangu's Atax result has a negative timing slack of $-0.30$\,ns, and the two models were evaluated under different search budgets and parallelism settings. Therefore, these results demonstrate the feasibility of using open-source LLMs for HLS optimization, rather than a strictly controlled comparison of model capabilities.



\section{Related Work}

\textbf{HLS Optimization.}
Existing HLS optimization methods mainly explore hardware directives
or apply compiler-based transformations. Directive-search approaches,
including AutoDSE~\cite{sohrabizadeh2022autodse},
COMBA~\cite{zhao2017comba}, and
FADO~\cite{du2023fado,du2024fado}, use heuristic search,
analytical models, or learned predictors to identify promising
configurations. ScaleHL-S~\cite{ye2022scalehls},
SEER~\cite{cheng2024seer}, and more recent holistic
frameworks~\cite{pouget2025unified,pouget2025holistic} further
optimize source code or intermediate representations and, in some
cases, combine code transformations with directive insertion.
Nevertheless, these methods generally operate within predefined
transformation spaces or fixed optimization orders, limiting their
ability to revisit source-level decisions using downstream feedback.

\textbf{LLM-Assisted HLS.}
Recent studies use LLMs for HLS code generation and optimization.
HLSPilot~\cite{xiong2024hlspilot} generates and refines synthesizable
designs, while HLSRewriter~\cite{xu2025hlsrewriter} focuses on
source-code refactoring. C2HLSC~\cite{collini2025c2hlsc} and
ChatHLS~\cite{li2025chathls} further demonstrate LLM-based
translation, analysis, and optimization for HLS. However, these approaches typically rely on one-shot, fixed-stage, or externally scripted workflows, with limited support for revising earlier
restructuring decisions or maintaining structured state across repeated
synthesis and rewriting iterations.



\section{Conclusion}
\label{sec:conslusion}

This paper presents HLSClaw, a skill-based agentic harness for HLS optimization that coordinates code restructuring, directive search, and synthesis feedback. HLSClaw achieves a 39.12$\times$ geometric-mean latency speedup with strong timing robustness, outperforming ScaleHLS, Claude Code, and Codex, while also supporting user-defined latency--resource trade-offs through goal-aware DSE.





\newpage
\balance
\bibliographystyle{ACM-Reference-Format}
\bibliography{ref}


\end{document}